\documentclass[draft]{agujournal2019}
\usepackage{url} 
\usepackage{lineno}
\usepackage[inline]{trackchanges} 
\usepackage{soul}
\usepackage{amsmath}
\usepackage{ragged2e}

\draftfalse

\journalname{Enter journal name here}

\begin{document}
\begin{justify}

%
%


\title{Data-driven Mesoscale Weather Forecasting Combining Swin-Unet and Diffusion Models}

%
%




\authors{Yuta Hirabayashi\affil{1}, Daisuke Matsuoka\affil{1}}


\affiliation{1}{Japan Agency for Marine-Earth Science and Technology(JAMSTEC), Yokohama, Japan}




\correspondingauthor{Yuta Hirabayashi}{yuta.hirabayashi@jamstec.go.jp}



\begin{keypoints}
\item A mesoscale weather prediction model that combines Swin-Unet is proposed as a deterministic Predictor and a diffusion model as a probabilistic Corrector.
\item Both the Predictor and Corrector were trained independently, enabling flexible updates to the deterministic Predictor without retraining the Corrector.
\item The proposed approach enhances high spatial frequency components, leading to improved accuracy in strong rainfall forecasts.
\end{keypoints}

%
%

%
%


\begin{abstract}
      Data-driven weather prediction models exhibit promising performance and advance continuously.
      In particular, diffusion models represent fine-scale details without spatial smoothing,
      which is crucial for mesoscale predictions, such as heavy rainfall forecasting.
      However, the applications of diffusion models to mesoscale prediction remain limited.
      To address this gap, this study proposes an architecture that combines a diffusion model with Swin-Unet as a deterministic model,
      achieving mesoscale predictions while maintaining flexibility. The proposed architecture trains the two models independently,
      allowing the diffusion model to remain unchanged when the deterministic model is updated.
      Comparisons using the Fractions Skill Score and power spectral analysis demonstrate that incorporating the diffusion model
      leads to improved accuracy compared to predictions without it. These findings underscore the potential of the proposed architecture
      to enhance mesoscale predictions, particularly for strong rainfall events, while maintaining flexibility.
\end{abstract}

\section*{Plain Language Summary}
Recent advances in artificial intelligence have led to data-driven weather prediction models that are comparable to conventional numerical weather prediction models.
Among these models, diffusion models are particularly promising due to their ability to generate realistic weather patterns while preserving fine details.
This ability makes them particularly useful for predicting mesoscale weather events, such as heavy rainfall. 
In this study, we propose an architecture for mesoscale weather prediction, integrating a Swin-Unet model for time evolution with a diffusion model for fine-scale correction.
By training the two models independently, our approach allows flexible updates to the prediction model without retraining the diffusion model. 
This approach enhances the representation of fine-scale structures, improving accuracy in predicting strong rainfall.


%
%

%


%
%
%
%

\section{Introduction}


Accurate  forecasting of mesoscale phenomena, including heavy rainfall, is essential.
Mesoscale refers to the scale between the synoptic ($\sim$1000 km, $\sim$1 day) and three-dimensional turbulent scales ($\sim$100 m, $\sim$1 min) \cite{Orlanski1975ARS,Craig2018}. 
Examples of mesoscale phenomena include fronts, squall lines, and mesoscale convective systems \cite{Craig2018}.
Mesoscale phenomena are often associated with heavy rainfall \cite{Houze2004, Schumacher2005}, which causes notable disasters.
Accurate phenomena forecasting helps mitigate disasters by informing people in advance.
To accurately forecast mesoscale phenomena, high-resolution numerical weather prediction (NWP) models play a crucial role.
For instance, the Japan Meteorological Agency operates regional mesoscale models designed to mitigate disasters caused by torrential rain \cite{Ishida2022}.
Mesoscale models numerically solve partial differential equations on high-resolution grids,
whereas the increasing spatial resolution of these models comes at the cost of considerable computational demands.

Data-driven weather prediction models are gaining increasing attention for their potential to reduce computational costs while archiving high accuracy.
These models use advanced machine learning algorithms trained on reanalysis datasets to simulate the temporal evolution of meteorological variables, 
leveraging graphics processing units (GPUs) or tensor processing units (TPUs) for enhanced computational efficiency \cite{Lam2023,Bi2023}. 
For example, GraphCast \cite{Lam2023} has outperformed the Integrated Forecasting System (IFS) 
across most variables in terms of the root mean squared error (RMSE), 
generating 10-day forecasts at $0.25^\circ$ resolution globally in under 1 min using a single TPU. 
In addition to average error metrics such as RMSE, other characteristics such as the physical realism of data-driven models have also been investigated \cite{Bonavita2024, Charlton-Perez2024, Ben-Bouallègue2024}.
For example, \citeA{Bonavita2024} reported that the spectra of forecasts from Pangu-weather \cite{Bi2023}, a data-driven weather forecasting model, 
exhibit reduced energy at wavelengths finer than $\sim$500–700 km.
This suggests that data-driven weather prediction models fail to represent fine-scale features, which are essential for mesoscale prediction.
This is partly because they are trained deterministically using mean squared error (MSE) as the loss function,
suggesting that a probabilistic approach could address this issue.

Diffusion models \cite{Ho2020,Song2021} have the potential to enable data-driven models to represent fine-scale features
due to their ability to generate predictions that follow the probabilistic distribution of a training dataset.
Diffusion models consist of two primary processes: diffusion, which transforms high-dimensional data into gaussian noise, and denoising, 
which generates samples by approximating the probabilistic distribution of data starting from Gaussian noise.
An approach to the application of diffusion models to weather forecasting is 
to generate meteorological variables conditioned on the variables at the previous time step, such as GenCast \cite{Price2025}.
Another approach is to use a diffusion model to correct the output of a deterministic machine learning model for time evolution, 
such as ArchesWeatherGen \cite{Couairon2024}.
The former approach is simple, whereas the latter approach has the advantages of 
enabling the diffusion model to converge with low computational costs during training
and allows the use of state-of-the-art deterministic models \cite{Couairon2024,Mardani2025,Finn2024}.
For example, ArchesWeatherGen, which uses a diffusion model with a transformer-based deterministic model,
has demonstrated that its power spectra are much closer to those of a reanalysis dataset
than those of the deterministic model without the diffusion model.
Although both GenCast and ArchesWeatherGen have limited capability to capture mesoscale information 
because they are trained on a reanalysis dataset at a global $0.25^\circ$ resolution, 
their demonstrated ability to represent fine-scale features suggests that diffusion models have the potential to be applicable to mesoscale prediction.

However, research on applying diffusion models to mesoscale prediction remains limited despite their potential. 
An exception is StormCast \cite{Pathak2024}, which consists of a combination of a diffusion model and a U-Net as a deterministic model, 
similar to ArchesWeatherGen.
StormCast has demonstrated their potential to forecast mesoscale phenomena, such as convective cluster evolution.
This suggests that combining diffusion and deterministic models is a promising approach for mesoscale prediction as well as global forecasting.
However, this combination lacks flexibility, as retraining the diffusion model is necessary whenever the deterministic model is adjusted.
This is because the diffusion model is trained on a paired dataset that consists of the deterministic model's output as well as the ground truth.
An architecture in which the diffusion model is independently trained on an unpaired dataset containing only ground truth
could provide flexibility and broad applicability for mesoscale weather prediction.

This study proposes an architecture that combines a transformer-based model as a deterministic Predictor and a diffusion model as a probabilistic Corrector 
to enable mesoscale predictions while preserving flexibility in model configuration and updates. 
The architecture builds on the approach described in \citeA{Bischoff2024}, 
which demonstrated that a diffusion model trained on an unpaired dataset produced promising results for downscaling tasks by converting low-resolution fields to high-resolution fields. 
This study extends their approach by treating the output of the deterministic Predictor as the low-resolution data in their study,
and also represents an extension of StormCast in terms of enhancing flexibility. 
This study aims to reveal whether this architecture can enhance the performance of mesoscale prediction while maintaining flexibility.

\section{Data and Methods}
\label{sec:method}

The data for this study was prepared by preprocessing and sampling from the regional climate downscaling dataset RCDSJRA-55 \cite{Kawase2023}. 
This dataset includes two types of resolutions: DS20km (20 km grid spacing) and DS5km (5 km grid spacing). 
DS5km was used in this study, incorporating five surface variables and five upper air variables as input features and ground truth, respectively.
The surface variables include temperature at 1.5 m above ground, u-component and v-component of winds at 10 m above ground, mean sea level pressure, and total precipitation. 
The upper air variables include temperature, u- and v-components of winds, geopotential height, and dew point depression at pressure levels of 200, 300, 500, 700, 850, 925, and 1000 hPa.
Data at 00, 06, 12, and 18 UTC were used for training, validation, and testing.
The central 512 × 768 section of the horizontal grid was sampled from the original dataset (527 × 804). 
An example of the data is shown in Figure \ref{fig:example}. 
All variables were normalized for weather prediction models.

\begin{figure}
      \noindent\includegraphics[width=\textwidth]{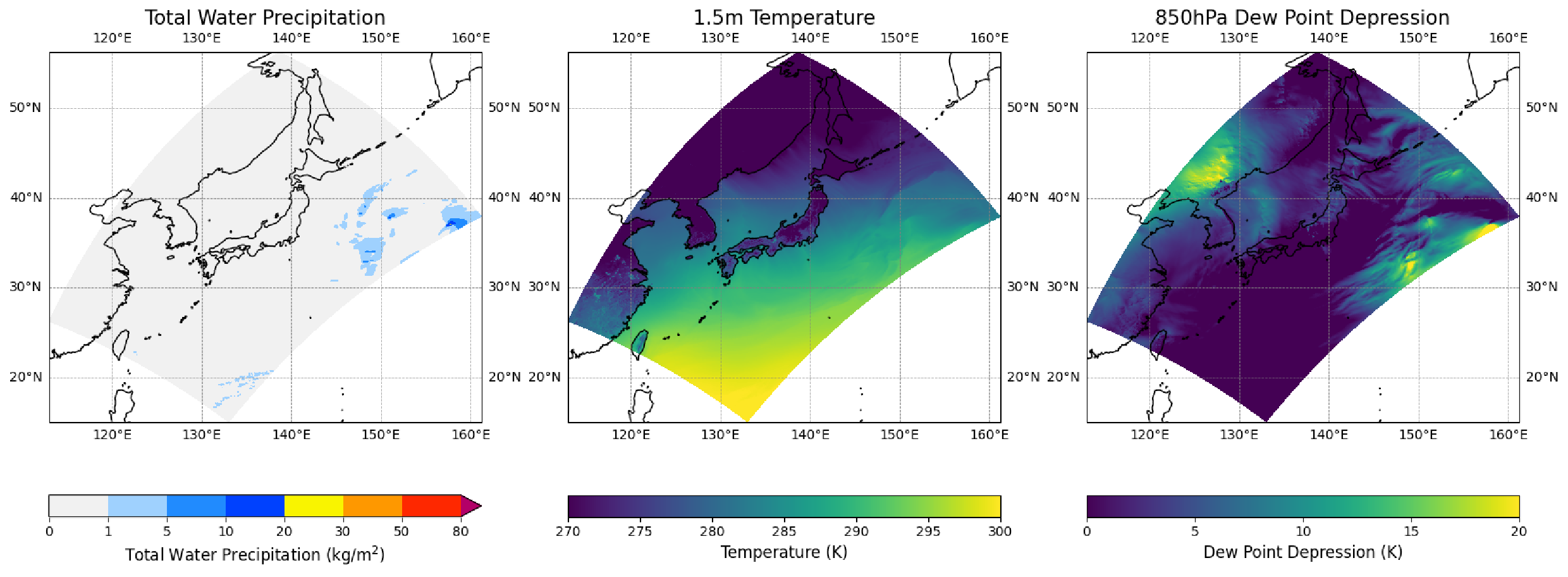}
      \caption{
            Examples of the reanalysis data used in this study(RCDSJRA-55), after sampling of the central 512 × 768 section of the horizontal grid.
            From left to right, the plots show total water precipitation, 
            1.5 m temperature, and 850 hPa dew point depression over the region of interest, including Japan and its surrounding areas.
      }
      \label{fig:example}
\end{figure}

Figure \ref{fig:architecture} illustrates the schematic design of our prediction method, 
which integrates a Swin-Unet model \cite{Cao2021, Fan2022} as a deterministic Predictor and a diffusion model as a probabilistic Corrector. 
The Swin-Unet predicts the expected meteorological variables at the next time step, whereas the diffusion model probabilistically adjusts residuals to enhance high spatial frequency structures.
The Swin-Unet uses a U-shaped architecture with Swin transformer-based blocks, an extension of Vision Transformer \cite{Dosovitskiy2020}. 
The Swin-Unet  uses SwinTransformer V2 \cite{Liu2021} with a base embedding size of 96 and a window size of 4. To prevent checkerboard artifacts, 
we adopted a dual up-sample block combining bilinear and subpixel up-sample methods \cite{Shi2016, Fan2022}. Other parameters were determined by following \citeA{Cao2021}.
In contrast , the diffusion model leverages stochastic differential equations (SDEs) to learn the probabilistic distribution of the dataset. 
Based on the U-Net architecture in \citeA{Dhariwal2021}, it includes 4 encoder and decoder layers with a base embedding size of 64, two residual blocks per resolution, and channel scaling factors [1, 2, 2, 2].
No attention blocks are applied, as in \citeA{Bischoff2024}. Training protocols and parameters were determined according to \citeA{Karras2022}, with adjustments to reduce training time and GPU memory usage.

\begin{figure}
      \noindent\includegraphics[width=\textwidth]{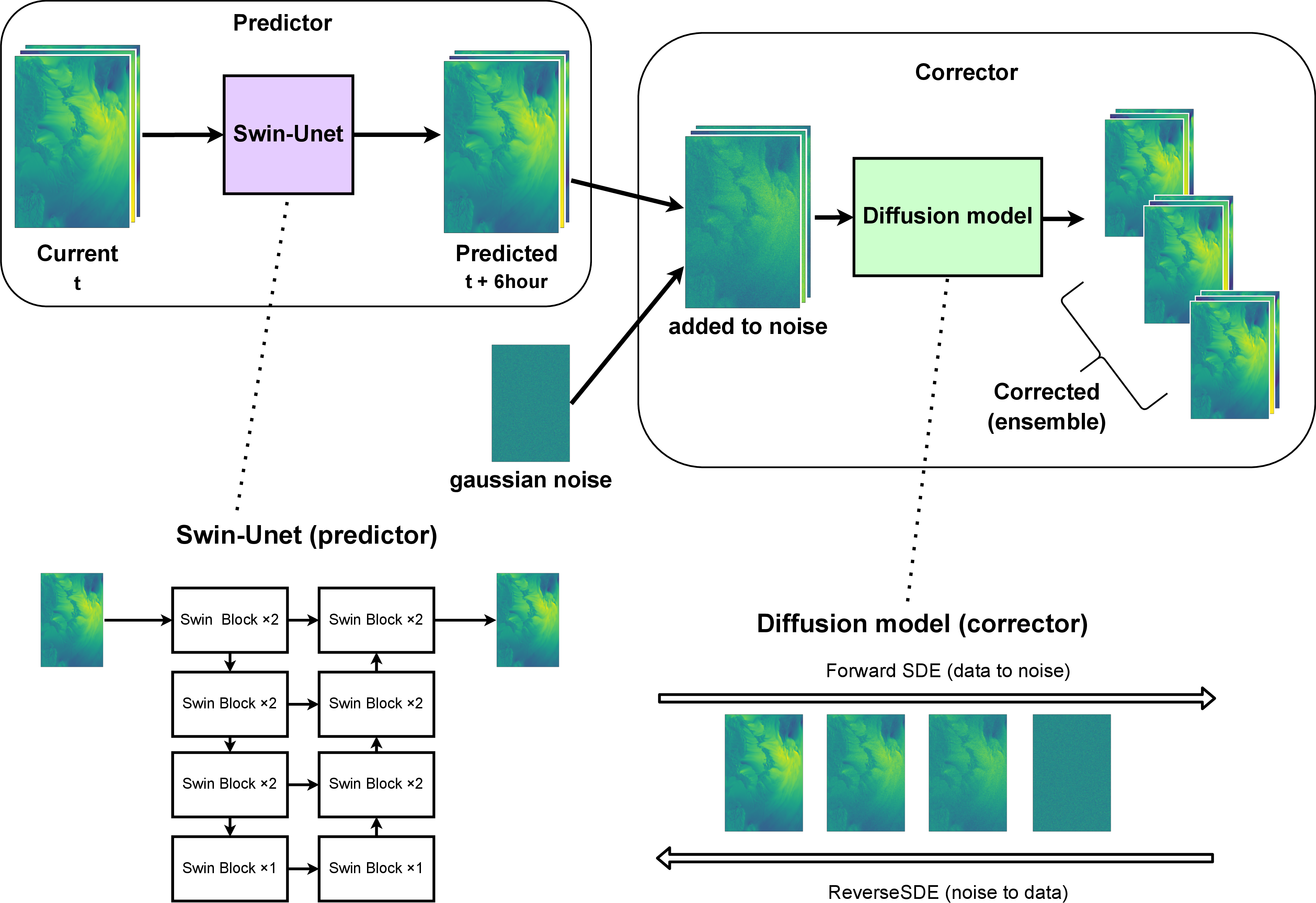}
      \caption{
            Schematic overview of the Predictor-Corrector framework combining Swin-Unet and diffusion models. 
            The Predictor (Swin-Unet) takes an input state at time t and predicts the state at t + 6 h, for a detailed diagram, see \citeA{Cao2021}, Fig. 1.
            Gaussian noise is added to the predicted state to apply the Corrector (diffusion model). 
            The Corrector refines the output of the Predictor added to the noise through a denoising process, generating an ensemble of corrected outputs. 
      }
      \label{fig:architecture}
\end{figure}

The deterministic Predictor and probabilistic Corrector were trained independently. 
The Predictor was trained on paired data samples with a 6 h interval to minimize the mean squared error, 
enabling it to predict 6 h ahead based on the current state. 
In contrast, the probabilistic Corrector was trained on unpaired data to approximate the probabilistic distribution of the dataset.
Both models were trained for 400 epochs with a batch size of 16, utilizing 8 NVIDIA A100 (40 GB) GPUs. 
The Predictor was optimized using Adam with a learning rate of $5 \times 10^{-4}$ and a weight decay of $3 \times 10^{-6}$, 
while the Corrector was optimized using AdamW with a learning rate of $1 \times 10^{-4}$, $\beta_1=0.9$, and $\beta_2=0.99$. 
Automatic mixed precision (AMP) was used during training for both models.

The prediction process involved generating the meteorological variables 6 h ahead using the Predictor, 
followed by obtaining a corrected ensemble forecast from the Corrector.  
The Corrector applied a reverse diffusion process to the Predictor's output with added Gaussian noise, 
following concepts from downscaling tasks \cite{Bischoff2024,Hess2024},  
where low-resolution data with added Gaussian noise was used as input.  
As described in \citeA{Bischoff2024}, the noise level \( \sigma\) was calculated as Equation (\ref{equation:PSD}) :
\begin{linenomath*}
      \begin{equation}  
            \sigma^2 = N \cdot \text{PSD}(k)
            \label{equation:PSD}
      \end{equation}
\end{linenomath*}
where \( k \) represents the wavenumber, \( N \) represents the grid size in the x-dimension, 
and PSD represents the power spectral density of each normalized variable,  
obtained by applying a Fourier transform in the x-direction and averaging in the y-direction.  
The wavenumber \( k \) was determined for each variable
as the point where the Predictor's power spectrum drops by a certain proportion of the true power spectrum.  
This proportion is a hyperparameter and was set to 0.1 based on the trade-off that a larger value limits the effect of the Corrector,  
whereas a smaller value leads to the collapse of lower-frequency structures in the Predictor \cite{Bischoff2024}.  
Subsequently, \( \sigma \) for each variable was optimized using Equation (\ref{equation:PSD}),
and the noise level \( \sigma \) used in the reverse diffusion process was determined as the median.    
This enables the lower-frequency structures (wavenumbers less than \( k \)) in the Predictor's output to be preserved,  
while high-frequency components are probabilistically predicted by the diffusion model. 
This noise level optimization was conducted using the validation dataset.  
Finally, an ensemble was generated by applying random noise variations with \( \sigma \) determined above.  
The ensemble size was set to 16, and each prediction was produced over 20 generation steps in the reverse diffusion process,  
similar to StormCast, which uses 18 steps.

Validating this method involved comparing the accuracy of the Predictor-Corrector to that of the Predictor alone, with both models trained on the same data. 
The training data consisted of 16 years (2000–2015), validation data covered 3 years (2016–2018), and test data consisted of 1 year (2019). 
After training, the noise level \( \sigma\) was optimized using the validation dataset, resulting in $\sigma = 2.148$,
which was applied during the test phase.
The performance test compared predicted values from both the Predictor-Corrector and the Predictor alone to the ground truth. 
For precipitation predictions, the fractions skill score (FSS), proposed by \citeA{Roberts2008}, was used to evaluate accuracy across various spatial scales. 
Although various versions of FSS exist, Equation (\ref{equation:FSS}) was used, as described in \citeA{Mittermaier2021},
where F and O are the spatial fractions for a given neighbourhood size n in the forecast and observed ﬁelds,
\(N_t\) is number of forecasts, \(N_x\) and \(N_y\) are number of grids in the x and y directions.
Following StormCast \cite{Pathak2024}, the Probability Matched Mean (PMM) \cite{Ebert2001} was also utilized to aggregate ensemble precipitation predictions. 
The detailed computation of PMM followed the description in Section 2.b of \citeA{Clark2017}. 
The FSS calculations for the Predictor-Corrector ensemble were conducted based on the post-PMM results.
For other variables, the ensemble mean RMSE for the Predictor-Corrector and the RMSE for the Predictor alone were calculated.
Power spectra were also computed for each variable to assess performance across spatial frequencies, using Fourier transforms in the x-direction and averaging in the y-direction. 
These metrics were calculated using the region shown in Figure \ref{fig:example}, excluding the upper and lower 5-grid boundaries in the y-direction due to noise patterns caused by the diffusion model. 

\begin{linenomath*}
      \begin{equation}
            \langle \text{FSS}_{(n)} \rangle =
            1 - \left\{
            \frac{
            \sum_{k=1}^{N_t}
            \left[
            \sum_{i=1}^{N_x} \sum_{j=1}^{N_y} 
            \left(F_{(n),ij} - O_{(n),ij}\right)^2
            \right]_k
            }{
            \sum_{k=1}^{N_t}
            \left[
            \sum_{i=1}^{N_x} \sum_{j=1}^{N_y} 
            F_{(n),ij}^2
            \right]_k
            + 
            \sum_{k=1}^{N_t}
            \left[
            \sum_{i=1}^{N_x} \sum_{j=1}^{N_y} 
            O_{(n),ij}^2
            \right]_k
            }
            \right\}
      \label{equation:FSS}
      \end{equation}
\end{linenomath*}

\section{Results}

Figure \ref{fig:example_of_results_rain} presents a case study comparing the models with and without the Corrector for precipitation, wind, and mixing ratio at 925 hPa during a heavy rainfall event. 
This event was influenced by a typhoon over the sea to east of Japan and a low-pressure system to south of Japan's main island.
The top panels of Figure \ref{fig:example_of_results_rain} shows that the Predictor-Corrector predicts maximum rainfall close to the ground truth.
Specifically, the Predictor-Corrector and Predictor alone predicted maximum rainfall of 63.1 and 55.6 \(\mathrm{kg \, m^{-2} \, h^{-1}}\), respectively, 
whereas the ground truth showed 83.5 \(\mathrm{kg \, m^{-2} \, h^{-1}}\).
The bottom panels illustrate wind convergence and significant vapor transport from the ocean in both models. 
Notably, the Predictor-Corrector demonstrates a more distinct convergence of the mixing ratio 
in the region indicated by the red ellipse in Figure \ref{fig:example_of_results_rain}
compared to the Predictor alone.

\begin{figure}
      \noindent\includegraphics[width=\textwidth]{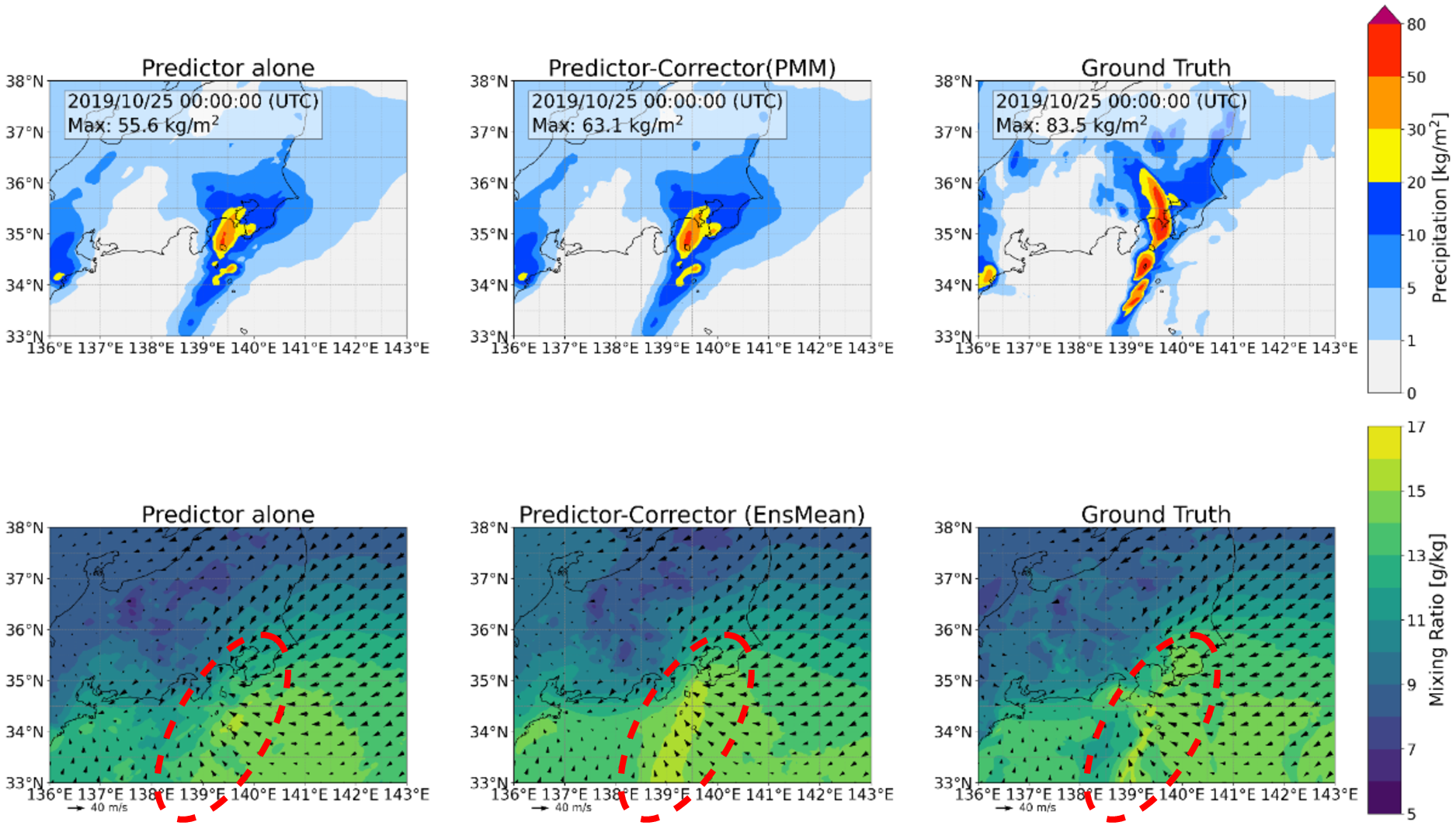}
      \caption{
            The top row shows a comparison of precipitation predictions and ground truth at 00:00 UTC on 25th October 2019, 
            whereas the bottom row provides a comparison of wind and mixing ratio at 925 hPa.
            In each row, the left column presents predictions from the Predictor alone model, 
            the middle column shows results from the Predictor-Corrector PMM model, and the right column displays the ground truth.
      }
      \label{fig:example_of_results_rain}
\end{figure}

Figure \ref{fig:example_of_results_weather} presents examples of meteorological variables predictions from the models 
with and without the Corrector at the same time as Figure \ref{fig:example_of_results_rain}. 
The Predictor-Corrector ensemble (Ens-Sample), a randomly selected member from the ensemble predictions generated by the Predictor-Corrector model,
represents more detailed structures compared to both the Predictor alone and the ensemble mean of the Predictor-Corrector (Ens-Mean), 
particularly for the u-component wind (1000 hPa), v-component wind (1000 hPa), and dew point depression (850 hPa).
For example, in the v-component wind, a pronounced strong feature is evident in both the Ens-Sample and the ground truth
in the region indicated by the red ellipse, 
although it appears spatially smoothed in the Predictor alone and Ens-Mean. 
The effect of the Corrector varies by variable; for instance, 
no significant differences were observed in temperature (500 hPa) or geopotential height (500 hPa), 
as they exhibit fewer fine-scale structures even in ground truth compared to the u-component wind (1000 hPa), v-component wind (1000 hPa), 
and dew point depression (850 hPa).

\begin{figure}
      \noindent\includegraphics[width=\textwidth]{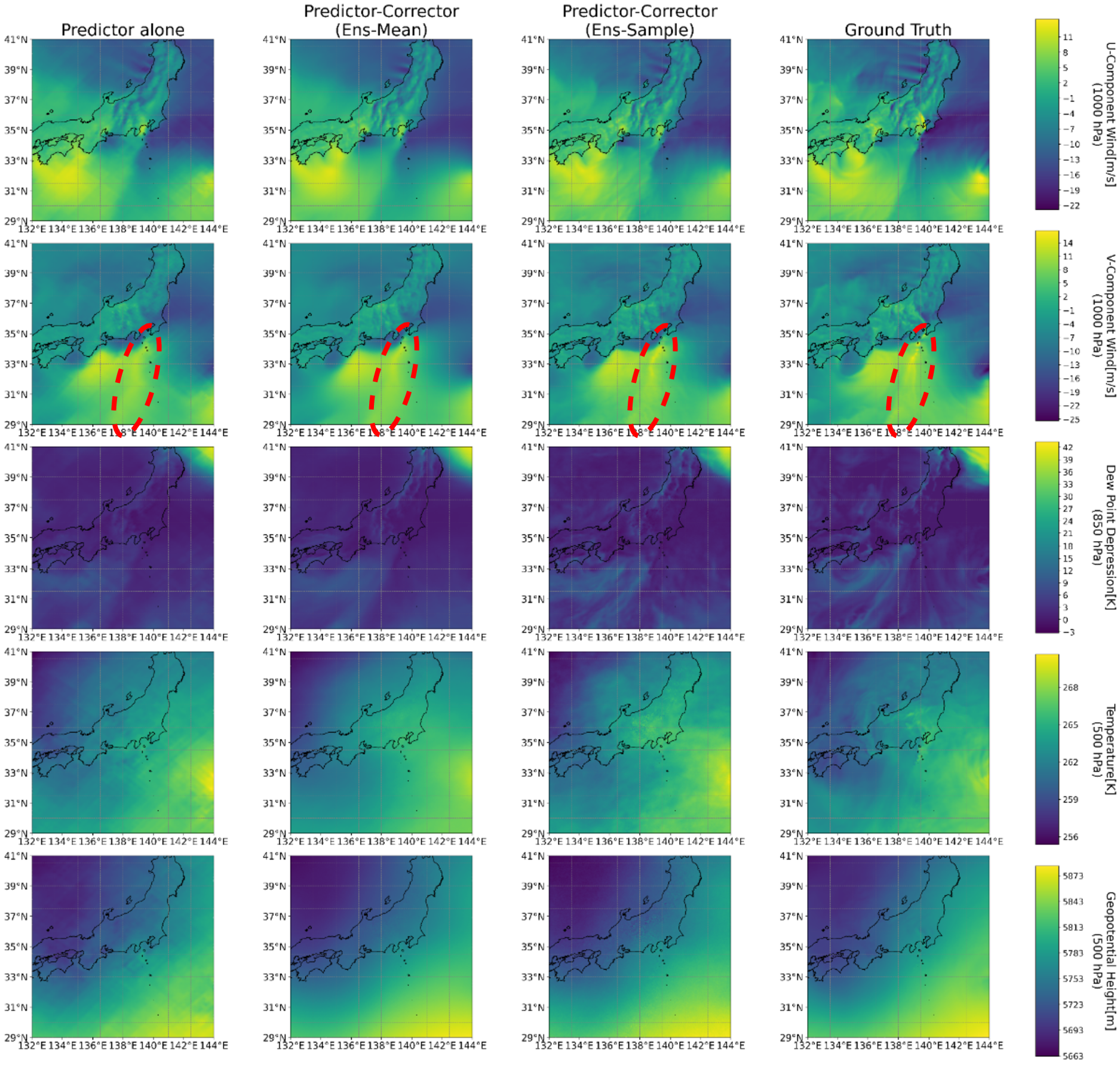}
      \caption{
            Comparison of meteorological variable predictions between the Predictor alone, Predictor-Corrector (Ens-Mean), Predictor-Corrector (Ens-Sample), and Ground Truth. 
            The columns represent different model outputs, with Ground Truth on the far right, followed by the Predictor-Corrector (Ens-Mean), Predictor-Corrector (Ens-Sample), and Predictor alone. 
            The rows represent different meteorological variables: u-component wind (top), v-component wind, dew point depression, temperature, and geopotential height (bottom). 
      }
      \label{fig:example_of_results_weather}
\end{figure}

FSS values across various spatial scales and thresholds were calculated to evaluate the skill of precipitation prediction.
Figure \ref{fig:FFS_total} illustrates the effect of the Corrector by comparing the FSS values for Predictor alone and Predictor-Corrector. 
The improvement is particularly notable at thresholds such as 20 \(\mathrm{kg \, m^{-2} \, h^{-1}}\) and 30 \(\mathrm{kg \, m^{-2} \, h^{-1}}\), 
where Predictor-Corrector consistently achieves high FSS values, especially at large spatial scales. 
In particular, at a spatial scale of 100 km, the FSS values were 0.200 and 0.248, respectively, indicating a 24 \% improvement by the Corrector.
This finding indicates enhanced predictive performance for strong precipitation events,
although the accuracy of the predicted location remains limited.

\begin{figure}
      \noindent\includegraphics[width=\textwidth]{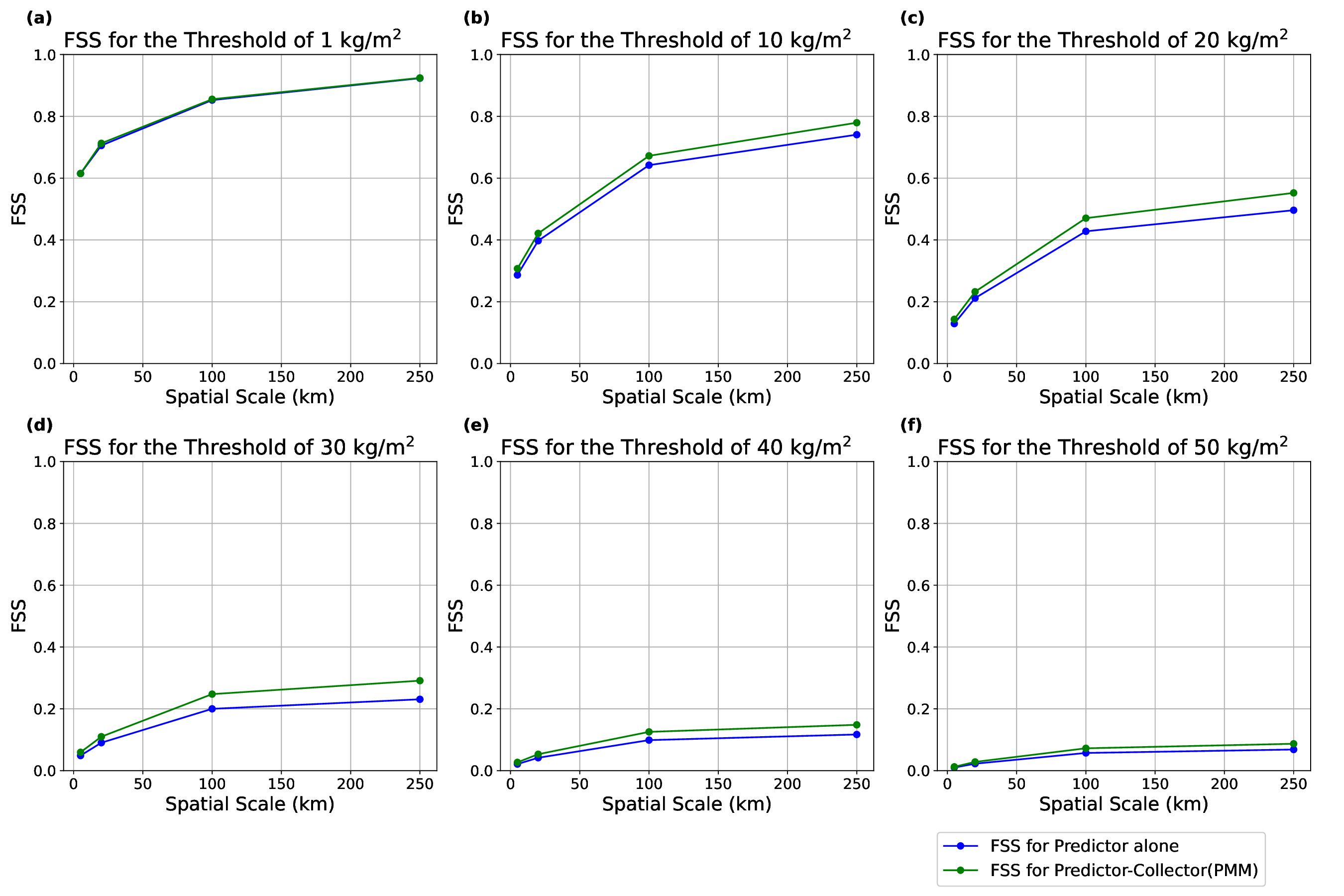}
      \caption{
            Fractions Skill Score (FSS) comparisons between the Predictor alone and Predictor-Corrector PMM models across different spatial scales and precipitation thresholds. 
            Panels (a) to (f) represent the FSS for increasing precipitation thresholds: 
            1, 10, 20, 30, 40, and 50 \(\mathrm{kg \, m^{-2} \, h^{-1}}\), respectively. 
            The horizontal axis shows the spatial scale in kilometers, while the vertical axis represents the FSS. 
      }
      \label{fig:FFS_total}
\end{figure}

Figure \ref{fig:FFT} illustrates the improvements in the power spectra of each variable achieved by using the Corrector. 
The Predictor-Corrector model shows better agreement with the ground truth compared to the Predictor alone, 
particularly for structures finer than 500 km, as indicated by the red lines in Figure \ref{fig:FFT}. 
While the Predictor alone exhibits reduced power for many variables at small scales, the Predictor-Corrector mitigates this attenuation effectively.
Significant improvements were observed for u- and v-component winds at the surface and 1000 hPa, as well as dew point depression at 1000 hPa. 
These findings align with the observations presented in Figure \ref{fig:example_of_results_weather}.

\begin{figure}
      \noindent\includegraphics[width=\textwidth]{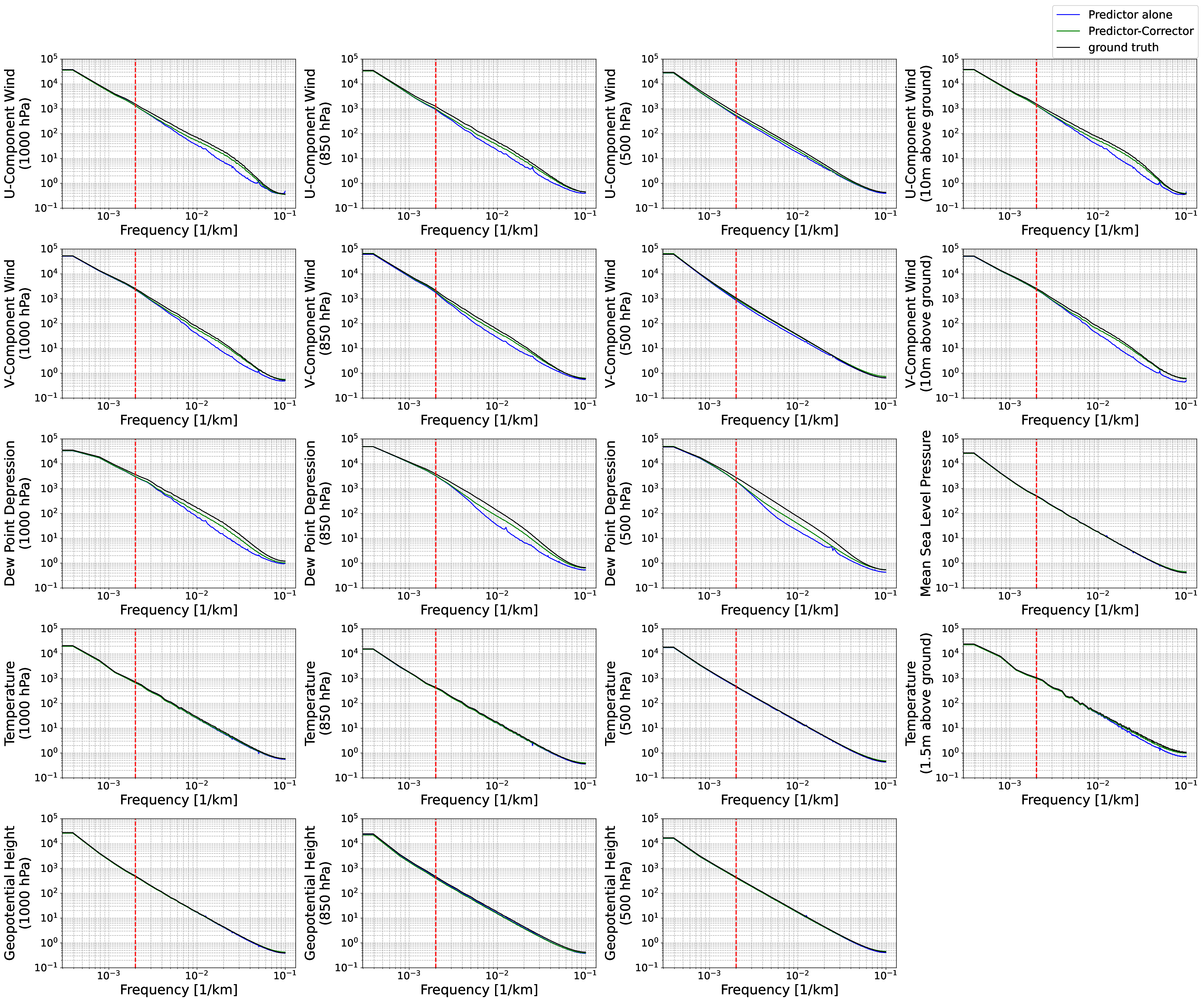}
      \caption{
            Power spectral density (PSD) plots for various atmospheric variables, obtained by applying a Fourier transform in the x-direction and averaging in the y-direction. 
            Each red line indicates the reference wavelength of \(2 \times 10^{-3}  [\mathrm{km^{-1}}]\), corresponding to a spatial scale of 500 km. 
            Each plot shows the PSD results derived from normalized data for three cases: Predictor alone, Predictor-Corrector, and ground truth data. From the top-left to bottom-right, 
            the variables displayed are U-component wind (1000 hPa, 850 hPa, 500 hPa, 10 m above ground), 
            v-component wind (1000 hPa, 850 hPa, 500 hPa, 10 m above ground), temperature (1000 hPa, 850 hPa, 500 hPa, 1.5 m above ground), 
            dew point depression (1000 hPa, 850 hPa, 500 hPa), geopotential height (1000 hPa, 850 hPa, 500 hPa), and mean sea level pressure. 
      }
      \label{fig:FFT}
\end{figure}

Figure \ref{fig:RMSE} compares RMSE between the Predictor-Corrector and Predictor alone. 
The results show that the ensemble mean RMSE of the Predictor-Corrector is comparable to the RMSE of the Predictor alone for u-component wind, v-component wind, and dew point depression at 925 hPa, 1000 hPa, and the surface. 
However, for other pressure levels and variables, the Predictor-Corrector demonstrates poorer performance.
These improvements in specific variables and pressure levels align with the frequency enhancement effects observed in Figure \ref{fig:FFT}, 
suggesting that the Predictor-Corrector is particularly effective where high-frequency components are significant.

\begin{figure}
      \noindent\includegraphics[width=\textwidth]{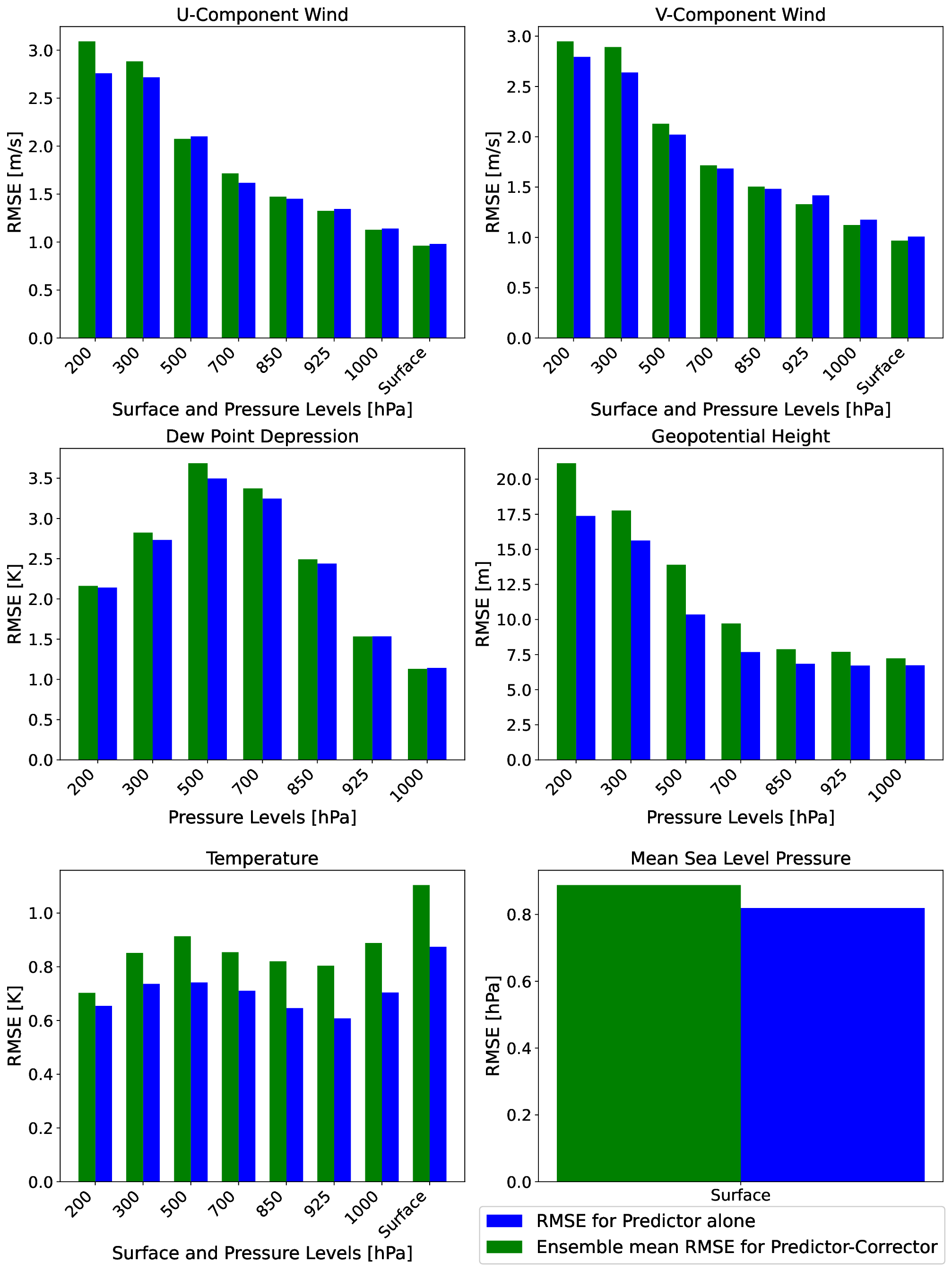}
      \caption{
            RMSE comparison between the Predictor alone and Ensemble mean for Predictor-Corrector models across various meteorological variables and pressure levels. 
            The six panels show RMSE for u-component wind, v-component wind, dew point depression, geopotential height, temperature, and mean sea level pressure (MSLP). 
            The horizontal axis represents surface and pressure levels in hectopascals (hPa), and the vertical axis shows RMSE in the respective units. 
            The blue bars correspond to the Predictor alone model, 
            whereas the green bars represent the Ensemble mean of the Predictor-Corrector model.
      }
      \label{fig:RMSE}
\end{figure}

\section{Discussion and Conclusions}

The improvements in FSS (Figure \ref{fig:FFS_total}) and power spectral analysis (Figure \ref{fig:FFT}) achieved by the Corrector indicate that it effectively enhances the effective resolution, 
leading to improved accuracy in rainfall prediction. 
The spatial smoothing observed in the Predictor alone aligns with findings from previous studies, 
such as the characteristics of data-driven models in global simulations \cite{Bonavita2024} and the underestimation of extreme event intensity reported by \citeA{Charlton-Perez2024}.
In contrast, the Corrector, utilizing a diffusion model, enhances high spatial frequencies and improves precipitation accuracy, 
particularly for strong rainfall events. 
This aligns with findings in physics-based NWP \cite{Kato2020}, where fine spatial resolution enhances predictions of intense rainfall. 
Our case study (Figure \ref{fig:example_of_results_rain}, Figure \ref{fig:example_of_results_weather}) and power spectral analysis (Figure \ref{fig:FFT}) further reveal that these improvements are most pronounced in lower atmospheric variables, 
such as wind and dew point depression, which are critical for predicting heavy rainfall.

Comparison of RMSE between the Predictor-Corrector and Predictor alone (Figure \ref{fig:RMSE}) reveals two key characteristics of the Corrector.
First, RMSE improvements are observed only in specific variables that lack high spatial frequency components in the Predictor’s results.
This suggests that the Corrector's effectiveness depends on the Predictor's accuracy for each variable. 
As noted in Section~\ref{sec:method}, the noise level, which determines the preserved spatial frequencies, is constant across all variables.
Consequently, the Corrector enhances high spatial frequency components for variables such as wind and dew point depression at lower levels, 
where these components are present in the ground truth.
This variable-dependent characteristic of the Corrector method aligns with findings by \citeA{Hess2024}, where diffusion models were used for downscaling. 
Thus, RMSE improvements are limited to such variables, whereas other variables exhibit poor performance in terms of RMSE.
Adjusting the noise level for each variable could help strike a balance between improving prediction accuracy and preserving the Predictor’s results.
Second, even for variables that show improvements in high spatial frequency components, RMSE improvements remain limited.
This is partly because the Corrector preserves the low spatial frequency components in the Predictor’s results, such as the shape or position of a heavy rainfall area.
Their preservation limits the overall RMSE improvement because these low spatial frequency components contribute significantly to the RMSE of the Predictor.

This study demonstrated that combining deterministic prediction models with diffusion models improves mesoscale weather prediction, particularly for strong rainfall.
Specifically, the combination of a transformer-based model (Swin-Unet) as a deterministic model and 
a diffusion model effectively enhances high spatial frequency components, leading to improved rainfall prediction accuracy.
A key advantage of this approach is its flexibility. As highlighted in Section~\ref{sec:method}, 
the models are trained independently, allowing updates of the deterministic model, such as retraining with new datasets or changing its architecture, 
without requiring retraining of the diffusion model. 
These updates are independent on the diffusion model, reducing computational costs when models are developed and trained.
While this study utilized the high-quality RCDSJRA-55 reanalysis dataset, further validation using actual observational data remains a crucial area for future work.

\clearpage

\acknowledgments
This work was supported by JST Moonshot R\&D (Grant Number JPMJMS2389)

%
%

\bibliography{export,manual}

%
%
%
%
%

\appendix

\end{justify}
\end{document}